\documentclass{article}
\usepackage{latexsym}
\usepackage{graphics}   
\usepackage{epsf}
\newlength{\largeurtexte}
\newcommand{\omitted}[1]{}
\newcommand{\cause}{\ensuremath{~causes~}}

\newcommand{\isa}{\ensuremath{\rightarrow_{IS-A}}}
\newcommand{\isaaux}{\ensuremath{\rightarrow_{IS-A(aug.)}}} 

\newcommand{\explicce}[2]{~\ensuremath \;\textit{explains}\;\;#1\; \textit{
    because\_possible }\{#2\}}
\newcommand{\explsfe}[2]{~\ensuremath \;\textit{explains}\;\;#1\; \textit{
    bec\_poss}\{#2\}}
\newcommand{\explicc}[2]{\ensuremath
  \;~\textit{explains}\;~#1\;~\textit{because\_possible}\;~#2 }
\newcommand{\explsf}[2]{\ensuremath \;\textit{explains}\;#1\;\textit{bec\_poss}\;#2 }
\newcommand{\explgf}[2]{\ensuremath \;\textit{explains}\;#1\;\textit{because\_possible}\;#2 }

\newenvironment{nonumberedmath}[1]{\begin{trivlist}\item[]\textbf{#1}\it}{\end{trivlist}}

\title{Ontology-based inference for causal explanation
  $^{*}$\footnotetext{$^{*}$ This preprint of a paper published in Integrated Computer-Aided Engineering Journal (publisher: IOS Press, editor:  Hojjat Adeli), 15(4), pp. 351--367, 2008)
is an extended version of \cite{BCM07}.}}
\author{Ph.\ Besnard\inst{1} \and M.-O. Cordier\inst{2} \and Y. Moinard\inst{2}}
\author{Ph.\ Besnard$^1$
\and  M.-O. Cordier$^{2\; 4}$
 \and Y. Moinard$^{3\; 4}$
}

\date{May 6, 2008}
\begin{document}
\maketitle
\addtocounter{footnote}{1}
\footnotetext{CNRS, IRIT, Universit\'{e} Paul Sabatier, 31062 Toulouse cedex,
  France, besnard@irit.fr}
\addtocounter{footnote}{1}
\footnotetext{Institution: Universit\'{e} de Rennes I}
\addtocounter{footnote}{1}
\footnotetext{Institution: INRIA}
\addtocounter{footnote}{1}
\footnotetext{Address: IRISA, Campus de Beaulieu, 35042 Rennes cedex, France, \{cordier,moinard\}@irisa.fr}
\addtocounter{footnote}{-4}

\bibliographystyle{alpha}

\begin{abstract}
We define an inference system to capture explanations based on causal statements, 
using an ontology in the form of an $IS$-$A$ hierarchy.
We first introduce a simple logical language which makes it possible to express that
a fact causes another fact and that a fact explains another fact.
We present a set of formal inference patterns from causal statements to explanation statements.
We introduce an elementary ontology which 
gives greater expressiveness to the system while staying close to propositional
reasoning.
We provide an inference system that captures the patterns discussed, 
firstly in a purely propositional framework, then in a datalog 
(limited predicate) framework.

\end{abstract}
\section{Introduction}
We are aiming at a logical formalization of explanations from causal statements. 
\mbox{For} example, 
it is usually admitted that fire is an explanation for smoke, on the grounds that fire causes smoke.
In other words, fire causes smoke is a premise from which it can be inferred that fire is an explanation for smoke.
In this particular example, concluding from cause to explanation is immediate 
but such is not always the case, far from it.
In general, the reasoning steps leading from cause to explanation are not so trivial:

\begin{nonumberedmath}{Example.}
We consider two causal statements: \\
\makebox[1.75em][r]{(i)}
Any ship that is about to sink causes her crew to launch some red rocket(s)\\
\makebox[1.75em][r]{(ii)}
On July the $14^{th}$, the celebration of the French national day causes the
launching of fireworks all over France.\\
So, if the place is a coastal city in France, on July the $14^{th}$, then red rockets being launched could be explained 
either by some ship(s) sinking or by a national day firework launched.
\end{nonumberedmath}
In this example, it is needed to acknowledge the fact that\\
\indent{\em a red rocket is a kind of (colourful) rocket}\\ 
in order to get the second explanation, which makes sense.

\begin{nonumberedmath}{Example (cont'd).}
Suppose that we now add the following statement: \\
\makebox[1.75em][r]{(i)}
Seeing a red rocket being launched triggers a rescue process.\\
Now, on July the  $14^{th}$ in a coastal city in France, a possible explanation for the triggering of the rescue process, as happens in practice, 
is that a 
national day firework has been launched.
\end{nonumberedmath}

Thus we say that {\em ``$\alpha$ explains $\beta$''} when adding $\alpha$ to our
knowledge, and using a ``suitable chain'' of causal and taxonomical 
information, $\beta$ is obtained. Which chains are ``suitable'' is one of the
subjects addressed in this text.
We define a dedicated inference system to capture explanations based 
on causal statements and stress that the r\^{o}le of ontology-based
information is essential. 
Our causal information is restricted to the cases where the causation 
never fails, rejecting e.g. ``smoking causes cancer''. 
We leave also for future work temporal aspects.
Also, we consider that
the causal information is provided by the user, we are not concerned by the
extraction of causal information as in scientific research. 
We provide a way to extract what we call explanations from 
causal (and ``ontological'') information given by the user:
we aim at providing all the 
(eventually tentative) explanations that can be obtained.
Then, some choice between these explanations should be made by the user,
depending of its needs, but this aspect is not considered here.

In the second section, we introduce the propositional logical language that we
propose to use, then we define the set of patterns dedicated to
inferring explanations from causal statements and ontological information.
In the third section we extend the formalism to a restricted predicate case
(a la ``datalog'', no quantifiers admitted in the formulas),
the ontology consisting in links between constant symbols.
We introduce two kinds of parameters for a predicate:
``existential'' and ``universal'' ones. Then we extend this to
predicates of any arity and we introduce also ontological links between
predicates.
In the fourth section, we discuss a few features of the formalism.
In the conclusion, we summarize the main points and we propose some possible
future work.

\section{The propositional formalism}\label{secprop}

\subsection{Vocabulary and first properties}\label{subvoc}

For the sake of clarity, we present the propositional version of the formalism 
first. We distinguish various types of statements in our formal system:
\begin{itemize}
\item[$C$:]
A theory expressing causal statements. E.g. $On\_alarm \cause
H\!eard\_bell$ 
or $Flu \cause Fever\_Temperature$.
\item[$O$:]
An ontology in the form of a set of $IS$-$A$ links between two items
which can appear in a causal statement.\\ E.g.,
$Temperature\_39 \isa Fever\_Temperature$,\\ 
$Temperature\_41 \isa Fever\_Temperature$, \\
$Heard\_loud\_bell \isa Heard\_bell$,\\
$Heard\_soft\_bell \isa Heard\_bell$.
\item[$W$:]
A classical propositional theory expressing truths 
(i.e., incompatible facts, co-occurring facts, $\ldots$). E.g.,
$Heard\_soft\_bell \rightarrow \neg Heard\_loud\_bell)$.
\end{itemize}
Intuitively, propositional symbols denote elementary properties
describing states of affairs, which can be ``facts'' or ``events''  
such as $Fever\_Temperature$,\\ $On\_alarm$, $Heard\_bell$.

The causal statements express causal relations between facts or events
expressed by these propositional symbols.
Some care is necessary when providing these causal and ontological atoms.
If ``$Flu \cause Fever\_Temperature$'', we will conclude
{\em $Flu$ explains $Temperature\_39$} from 
$Temperature\_39 \isa Fever\_Temperature$, but we cannot 
state $Flu \cause Temperature\_39$: we require that the causal information is
provided ``on the right level'' and in this case, $Temperature\_39$ is not 
on the right level.

Besides, our restricted ontology could be termed ``taxonomy''.

The formal system we introduce below is meant to infer,
from such premises $C\cup O \cup W$, formulas denoting explanations.
This inference will be denoted $\vdash_C$. 
The ontological  atoms express some common sense knowledge which is necessary
to infer these  ``explanations''.
Notice that a feature of our formalism 
is that standard implication alone cannot help to infer explanations
\cite{BCM06C,BCM07}.

In this section, $\alpha,\beta, \ldots$ denote the propositional 
atoms and $\Phi, \Psi, \ldots$ denote sets thereof. 
\begin{flushleft}\textbf{Atoms}\end{flushleft}
\begin{enumerate}
\item \emph{Propositional atoms}: $\alpha,\beta, \ldots$.
\item \emph{Causal atoms}: $\alpha \cause \beta$.
\item \emph{Ontological atoms}: $\alpha \isa \beta$.
\item \emph{Explanation atoms}: $\alpha \explicc{\beta}{\Phi}$.
\end{enumerate}
An ontological atom reads: $\alpha\mbox{ is a } \beta$.\\
An explanation atom reads: \emph{$\alpha$ is an explanation for $\beta$ because $\Phi$ is possible.}
\begin{flushleft}
\emph{Notation:} In order to help reading long formulas, explanation atoms are sometimes abbreviated as $\alpha \explsf{\beta}{\Phi}$.\\
\end{flushleft}
\textbf{Formulas}\nopagebreak
\begin{enumerate}
  \item \emph{Propositional formulas}: Boolean combinations of propositional
        atoms. 
  \item \emph{Causal formulas}: 
         Boolean combinations of causal or propositional atoms.
\end{enumerate}
The premises of the inference $\vdash_C$, namely $C \cup O \cup W$,
consist of propositional and causal formulas, and
ontological \emph{atoms} (no ontological formula). Notice that explanation
atoms cannot occur in the premises.

The properties of causal and ontological formulas we consider are as follows.
\begin{enumerate}
\item\label{propcausprop0} \textbf{Properties of the causal operator} 
  \begin{enumerate}
  \item \label{proofimplprop0} \emph{Entailing [standard] implication}:
      If $\alpha \cause \beta$, then $\alpha \rightarrow \beta.$
  \end{enumerate}
\item \label{propontprop0} \textbf{Properties of the ontological operator}
  \begin{enumerate}
  \item \label{ontoimplprop0} \emph{Entailing implication}:
      If $\alpha \isa \beta$, then $\alpha \rightarrow \beta.$
  \item\label{ontotransprop0} \emph{Transitivity}:
     If $a \isa b$ and $b \isa c$, then $a \isa c$.
  \item\label{ontorefprop0} \emph{Reflexivity}:
     $c \isa c$.
  \end{enumerate}
\end{enumerate}

Reflexivity is an unconventional property for an $IS$-$A$ hierarchy.
It is included here because it helps keeping the number of inference schemes low (see later).

$W$ is supposed to include (whether explicitly or via inference) all
the implications induced by the ontological atoms. 
For example, 
if $Heard\_loud\_bell \isa Heard\_bell$ is in $O$ then 
$Heard\_loud\_bell \rightarrow Heard\_bell$ is in $W$.
Similarly, $W$ is supposed to include all conditionals induced by the causal
statements in $C$. For example,  
if $Flu \cause Fever\_Temperature$ is in $C$, then $Flu \rightarrow Fever\_Temperature$ is \mbox{in $W$}.

\subsection{Patterns for inferring explanations}\label{subexplic}

A set of patterns, introduced in \cite{BCM07}, is proposed to infer
explanations from premises $C \cup O \cup W$.
Before providing the rules (see \S~\ref{proofsystem} below), let us 
motivate these rules by listing the main patterns that we consider as
desirable. The base case explains $\beta$ from $\alpha$ whenever $\alpha
\cause \beta$ (\S~\ref{subbasecase}). 
More elaborate cases explain $\beta$ from $\alpha$ 
whenever $\alpha \cause$ some $\beta'$ ontologically related with $\beta$
(\S~\ref{subexplicup}, \ref{subexplicdn}).
Finally, explanations should be transitive (almost) (\S~\ref{subtrans}).

\subsubsection{The base case}\label{subbasecase}

A basic idea is that what causes an effect can always be suggested as an
explanation when the effect happens to be the case: \\

\noindent\fbox{\parbox{\largeurtexte}{$
    \left\{\begin{array}[c]{ll}
    \mbox{If} & \alpha \cause \beta \\  \mbox{and} & W \not\models \neg \alpha
    \end{array}\right\}$ \hspace{5mm} then \hspace{5mm} 
    $\alpha \explicce{\beta}{\alpha}$.}}

\begin{nonumberedmath}{Example.}
Consider a causal model such that $W \not\vdash \neg Flu$ and $O$ is empty
whereas \hspace{20mm}$C = \{ Flu \cause Fever\_Temperature \}.$

Then, the atom \hfill $Flu \explicce{Fever\_Temperature}{Flu}$\\
is inferred.
That is, $Flu$ is an explanation for $Fever\_Temperature$.
\end{nonumberedmath}

Notice that since $Flu \rightarrow Fever\_Temperature$ is in $W$, we get in
fact that $W \not\vdash \neg Flu$ is equivalent to 
$W \not\vdash \neg (Flu \wedge Fever\_Temperature)$ which is why
$Fever\_Temperature$ is not included in the set of the ``conditions'' for this
explanation.

By the way, ``is an explanation'' must be understood as provisional.
Inferring that $Flu$ is an explanation for $Fever\_Temperature$ is a tentative
conclusion: Should $Flu$ be ruled out, e.g., $\neg Flu \in W$, then $Flu$ is
no longer an explanation for $Fever\_Temperature$.\\ 
Formally,
 with $Form=Flu \explsf{Fever\_Temperature}{\{Flu\}}$,\\ we get \hspace{1em}
$C \cup O \cup W \vdash_C Form$; \hspace{1em} but\hspace{1em}
$C \cup O \cup W \cup \{\neg Flu\} \not \vdash_C Form$.

\subsubsection{Wandering the IS-A hierarchy: Going upward}
\label{subexplicup}

What causes an effect can be suggested as an explanation for any consistent
ontological generalization of the effect:\\

\noindent\fbox{\parbox{\largeurtexte}{$
    \left\{\begin{array}[c]{lll}
     \mbox{If} &  \alpha \cause \beta,\\ &  \beta \isa \gamma,\\
     \mbox{and} & W \not\models \neg\alpha \end{array}\right\}$
      \hspace{5mm} then \hspace{5mm} $\alpha \explicce{\gamma}{\alpha}$.}}

\begin{nonumberedmath}{Example.}
  \begin{array}[t]{lll}
  C &= &\{ On\_alarm \cause Heard\_bell \}\\ 
  O &= &\{ Heard\_bell \isa Heard\_noise\}
  \end{array}\\

$O$ states that hearing a bell is more precise than hearing a noise.
Since $ On\_alarm$ is an explanation for $Heard\_bell$ from the base case, it
is also an explanation for $Heard\_noise$. Let the causal theory $CT$ consist
in the two preceding atoms of $C$ and $O$, 
$W$ containing nothing else than the implications induced by $C$ and $O$, that is:
\[W = \left\{\begin{array}{c}
On\_alarm \rightarrow Heard\_bell, \\
Heard\_bell \rightarrow Heard\_noise\\
\end{array}\right\}\]

We get $CT \vdash_C On\_alarm  \explsfe{Heard\_noise}{On\_alarm}$.\\

Then, we additionally know that 
hearing a fog-horn is more precise than hearing a noise, 
that a fog-horn is heard,
and that hearing a fog-horn is not hearing a bell.
This is expressed by the causal theory $CT'$, defined by the sets $C$ as
above, $O\cup O'$ and $W\cup W'$ with $O'$ and $W'$ as follows
($W'$ contains the new implication induced by $O'$, plus
the other additional information):

\[O'=\{Heard\_fog\textrm{-}horn \isa Heard\_noise\}.\]
\[W' = \left\{\begin{array}{c}
Heard\_fog\textrm{-}horn \rightarrow Heard\_noise,\\
Heard\_fog\textrm{-}horn,\\
\neg (Heard\_bell \leftrightarrow Heard\_fog\textrm{-}horn)
\end{array}\right\}\]
Even taking into account the fact that $Heard\_bell$ is an instance of
$Heard\_noise$, it can no longer be inferred that $On\_alarm$ is an
explanation for $Heard\_noise$:

 $CT' \not \vdash_C On\_alarm  \explsfe{Heard\_noise}{On\_alarm}$.

The inference fails because it would need $Heard\_noise$ to be of the $Heard\_bell$ kind (which is false, cf $Heard\_fog\textrm{-}horn$).
Technically, the inference fails because $W \cup W'\vdash \neg On\_alarm$.
\end{nonumberedmath}

\noindent%
The next example illustrates why resorting to ontological information is
essential when attempting to infer explanations: 
the patterns in the present \S~\ref{subexplicup} as well as in following
\S~\ref{subexplicdn} extend the base 
case for explanations  to \emph{ontology-based} consequences, not to any
consequences. 

\begin{nonumberedmath}{Example.}
Rain makes me growl. Trivially, I growl only if I am alive.
However, rain cannot be taken as an explanation for the fact that I am alive.

$C = \{Rain \cause I\_growl\}$, \hfill $O = \emptyset$, \hfill
$W = \{ I\_growl \rightarrow I\_am\_alive \}$.

\noindent We get: \hfill $C \cup O \cup W \not\vdash_C \hspace{1em}
Rain \explicce{I\_am\_alive}{Rain}$
\end{nonumberedmath}

\subsubsection{Wandering the IS-A hierarchy: Going downward}
\label{subexplicdn}

What causes an effect can presumably be suggested as an explanation when the
effect takes place in one of its specialized forms:\\

\noindent\fbox{\parbox{\largeurtexte}{$
\left\{\begin{array}[c]{ll}
 \mbox{ If} & \alpha \cause \beta,\\ & \gamma \isa \beta,\\ \mbox{and} &
W \not\models \neg (\alpha \wedge \gamma), \end{array}\right\}$ \hspace{3mm} then
\hspace{3mm} $\alpha \explicce{\gamma}{\alpha,\gamma}.$}}

\begin{nonumberedmath}{Example.}
Consider a causal model with $C$ and $O$ as follows:

$C = \{ On\_alarm \cause Heard\_bell \}$
\makebox[3em]{and}

$O = \left\{\begin{array}{c}Heard\_loud\_bell \isa Heard\_bell\\Heard\_soft\_bell \isa Heard\_bell\end{array}\right\}$\\

$O$ means that $Heard\_loud\_bell$ and $Heard\_soft\_bell$ are more precise than $Heard\_bell$.\\
Since $On\_alarm$ is an explanation for $Heard\_bell$,\,it also is an explanation for $Heard\_loud\_bell$ and similarly $Heard\_soft\_bell$.
This holds inasmuch as there is no statement to the contrary: 

The latter inference would not be drawn if for instance $\neg Heard\_soft\_bell$ 
or $\neg (Heard\_soft\_bell \wedge On\_alarm)$ 
were in $W$.\\
Formally, with $(Form\_loud)=$\\
$On\_alarm\explsf{Heard\_loud\_bell}{\{On\_alarm,Heard\_loud\_bell\}}$\\ and 
$(Form\_soft)=\\
On\_alarm \explsf{Heard\_soft\_bell}{\{On\_alarm,Heard\_soft\_bell\}}$:\\

\noindent$C \cup O \cup W \hspace{.5em}\vdash_C \hspace{.5em}(Form\_loud)$,
\hspace{2em} 
$C \cup O \cup W \hspace{.5em}\vdash_C \hspace{.5em}Form\_soft$, \hspace{2em}
and\\ 
$C \cup O \cup W \cup \{\neg ( Heard\_soft\_bell \wedge On\_alarm )\}
\hspace{.5em}\not\vdash_C \hspace{.5em}(Form\_soft)$
\end{nonumberedmath}

Here there is another example
\begin{nonumberedmath}{Example.}
$C = \{ Flu \cause Fever\_Temperature \}$ and\\
\makebox[4.6em]{} $O = \{Temperature\_39 \isa Fever\_Temperature\}$.
\\
$W$ contains no statement apart from those induced by $C$ and $O$, that is:
\[W=\{ Flu \rightarrow Fever\_Temperature,  Temperature\_39 \rightarrow Fever\_Temperature \}\]
Inasmuch as $Fever\_Temperature$ could be $Temperature\_39$, $Flu$ then counts as an explanation for $Temperature\_39$.
\[C \cup O \cup W \vdash_C 
Flu \explsf{Temperature\_39}{\{Flu,Temperature\_39\}}\]
Again, it would take $Flu \wedge Temperature\_39$ to be ruled out for the inference to be prevented.
\end{nonumberedmath}

\subsubsection{Transitivity of explanations}\label{subtrans}

We make no assumption as to whether the causal operator is transitive (from
$\alpha \cause \beta$ and $\beta \cause \gamma$ does $\alpha \cause \gamma$
follow?). 
However, we do regard inference of explanations as transitive which, in the
simplest case, means that if $\alpha$ explains $\beta$ and $\beta$ explains
$\gamma$ then $\alpha$ explains $\gamma$. 
Notice already that, since this transitivity of explanations ``gathers the
conditions'' 
(Point  \ref{transexplicprop5} \S~\ref{proofsystem} below), 
it is not absolute, and can easily be blocked.

The general pattern for transitivity of explanations takes two causal
statements, $\alpha \cause \beta_1$ and $\beta_2 \cause \gamma$ where $\beta_1$ and
$\beta_2$ are ontologically related, as premises in order to infer that
$\alpha$ is an explanation for $\gamma$. 

In the first form of transitivity, $\beta_2$ is inherited from $\beta_1$ by going
upward in the $IS$-$A$ hierarchy. \\ 

\noindent\fbox{\parbox{\largeurtexte}{$\left\{
  \begin{array}[c]{ll}
  \mbox{If} &\alpha \cause \beta_1,\;\;\beta_2 \cause \gamma,\\
   & \beta_1 \isa \beta_2,\\
  \mbox{and} & W \not\models \neg \alpha,
  \end{array}\right\}$ \hspace{1mm} then \hspace{2mm}
   $\alpha \explsfe{\gamma}{\alpha}$.}}

\begin{nonumberedmath}{Example.}
Sunshine makes me happy.
Being happy is why I sing.
Therefore, sunshine is a plausible explanation for the case that I am singing.
\[\begin{array}{@{}l}
C = \left\{\begin{array}{c}
Sunshine \cause I\_am\_happy \\
I\_am\_happy \cause I\_am\_singing 
\end{array}\right\}\\
W =  \left\{\begin{array}{c}
Sunshine \rightarrow I\_am\_happy \\
I\_am\_happy \rightarrow I\_am\_singing 
\end{array}\right\}
\end{array}\]
So, for the inference relation $\vdash_C$, $C \cup O \cup W$ infers the atom:
\[Sunshine \explicce{I\_am\_singing}{Sunshine}.\]
\end{nonumberedmath}

The above example exhibits transitivity of explanations for the simplest case that $\beta_1 = \beta_2$ 
in the pattern $\alpha \cause \beta_1$ and $\beta_1 \cause \gamma$ entail\\
 $\alpha \explicce{\gamma}{\alpha}$ 
(trivially, if $\beta_1 = \beta_2$ then $\beta_1 \isa \beta_2$).
This is one illustration that using reflexivity in the ontology relieves us from the burden 
of tailoring definitions to capture formal degenerate cases.

The next example exhibits the general case  $\beta_1 \not= \beta_2$
in the pattern given above. 
\begin{nonumberedmath}{Example.}
Let $O = \{Heard\_bell \isa Heard\_noise\}$ and
\[\displaystyle C = \left\{
\begin{array}{c}
On\_alarm \cause Heard\_bell\\
Heard\_noise \cause Disturbance
\end{array}\right\}\]
$W$ states the facts 
 induced by $C$ and $O$, that is:
\[W = \left\{\begin{array}[c]{c}
On\_alarm \rightarrow Heard\_bell \\
Heard\_noise \rightarrow Disturbance \\
Heard\_bell \rightarrow Heard\_noise
\end{array}\right\}\]
So, for the inference relation $\vdash_C$, $C \cup O \cup W$ infers the atom:
\[On\_alarm \explsf{Disturbance}{\{On\_alarm\}}.\]
\end{nonumberedmath}

In the second form of transitivity, $\beta_1$ is inherited from $\beta_2$ by
going downward in the $IS$-$A$ hierarchy.\\

\noindent\fbox{\parbox{\largeurtexte}{$
 \left\{\begin{array}[c]{ll}
 \mbox{If}&\alpha \cause \beta_1, \;\; \beta_2 \cause \gamma,\\
  & \beta_2 \isa \beta_1,\\
  \mbox{and}&W\not\models \neg (\alpha \wedge \beta_2),
 \end{array}\right\}$%
 \hspace{1mm}\mbox{then}\hspace{2mm} 
 $\alpha \explsfe{\gamma}{\alpha,\beta_2}.$}}

\begin{nonumberedmath}{Example.}
$O = \{Heard\_loud\_bell \isa Heard\_bell\}$ 
\[C = \left\{\begin{array}{c}
On\_alarm \cause Heard\_{bell}\\
Heard\_loud\_bell \cause Deafening
\end{array}\right\}\]
\[W = \left\{\begin{array}{c}
Heard\_loud\_bell \rightarrow Heard\_bell \\
On\_alarm \rightarrow Heard\_bell \\
Heard\_loud\_bell \rightarrow Heard\_Deafening
\end{array}\right\}\]
$On\_alarm$ does not \textbf{cause} $Heard\_loud\_bell$ (neither does it
\textbf{cause} $Deafening$),  
but it is an \textbf{explanation} for $Heard\_loud\_bell$ by virtue of the
downward scheme. 
Due to the base case, $Heard\_loud\_bell$ is in turn an explanation for
$Deafening$. 
In fact, $On\_alarm$ is an explanation for $Deafening$ by virtue of
transitivity. 
\end{nonumberedmath}

Considering a causal operator which is transitive would give the same
explanations  
but is obviously more restrictive as we may not want to endorse an account of 
causality which is transitive.
Moreover, transitivity for explanations not only seems right in 
itself but it also means that our model of explanations can be plugged with 
any causal system whether transitive or not.

The preceding examples are here to introduce the general pattern for
transitivity of explanations which is as follows:\\

\noindent\fbox{\parbox{\largeurtexte}{
$\left\{\begin{array}[c]{ll}
\mbox{If}&\alpha \explsf{\beta}{\Phi},\\&\beta \explsf{\gamma}{\Psi},\\
\mbox{and}&W\not\models \neg \bigwedge(\Phi \cup \Psi), 
\end{array}\right\}$\hspace{3mm}
\mbox{then} \hspace{5mm}
$\alpha \explsf{\gamma}{\Phi \cup \Psi}.$}}

\subsubsection{Explanation provisos and their simplifications}\label{ssubexpl}

Explanation atoms are written \hfill
$\alpha \explicc{\beta}{\Phi}$ \hfill \makebox{} 
as the definition is intended to make the atom true just in case it is successfully checked that the proviso is possible: 
An explanation atom is not to be interpreted as a kind of conditional statement.
Indeed, we do not write \emph{``if\_possible''}.
The argument in \emph{``because\_possible''} gathers those conditions that must be possible together if
$\alpha$ is to explain $\beta$ (there can be others: $\alpha$ can also be an
explanation of $\beta$ with respect to another set of arguments in
{\emph{``because\_possible''}}). 

Notice that the set of conditions in an explanation atom can often be
simplified.
Using $\bigwedge\Phi$ to denote the conjunction of the formulas in the set $\Phi$, 
the following general scheme 
amounts to simplifying the proviso attached to an explanation atom.\\

\fbox{\parbox{\largeurtexte}{
    \begin{center}$\left\{\begin{array}[c]{ll}\mbox{If}& 
\mbox{for all $i \in \{1,\cdots,n\}$, } 
\mbox{$\;\alpha \explgf{\beta}{(\Phi_i \cup \Phi)}$},\\
\mbox{and}&
W \models \bigwedge\Phi \rightarrow \bigvee_{i=1}^n \bigwedge\Phi_i,
\end{array}\right\}$\end{center}

\makebox[5mm]{} then \hspace{10mm}
$\alpha \explgf{\beta}{\Phi}.$}}
\\

As a simple motivating example for this general scheme, let us consider the
case where we have
\[\alpha \cause \beta \hspace{1em} and \hspace{1em} \beta \cause \gamma.\]

Then we get $\alpha \explsfe{\beta}{\alpha}$ and
$\beta \explsfe{\gamma}{\beta}$ by \S~\ref{subbasecase}, thus 
$\alpha \explsfe{\gamma}{\alpha,\beta}$ by the general pattern for
transitivity \S~\ref{subtrans}.
Now, we get $\alpha \rightarrow \beta$ from $\alpha \cause \beta$, thus,
$W \models \alpha$ is equivalent to $W \models \alpha \wedge \beta$:
the two sets of condition $\{\alpha\}$ and
$\{\alpha,\beta\}$ are equivalent here, which justifies to simplify the
explanation atom into
$\alpha \explicce{\gamma}{\alpha}$.
This provides a justification for the general pattern of simplification of the
set of condition where $n=1, \Phi=\{\alpha\}$ and $\Phi_1=\{\beta\}$. 

More complex examples may involve disjunctions, such as in the small following
example:

Let us suppose that we have derived the following two explanation atoms:\\
$\alpha \explsfe{\gamma}{\alpha,\beta_1}$
and $\alpha \explsfe{\gamma}{\alpha,\beta_2}$.
Let us suppose that $W$ contains $\alpha \rightarrow (\beta_1 \vee \beta_2)$.
Then, as soon as $\alpha$, together with either $\beta_1$ or $\beta_2$, is 
possible, we get that $\alpha$ explains $\gamma$.
Now, [$W \not \models \neg (\alpha \wedge \beta_1)$ or $W \not \models
\neg (\alpha \wedge\beta_2)$] is equivalent to 
[$W \not \models \neg (\alpha \wedge (\beta_1 \vee \beta_2))$]  and, 
from $\alpha \rightarrow (\beta_1 \vee \beta_2)$, this is equivalent to
[$W \not \models \neg \alpha$].

Thus, it is natural to get the explanation atom 
$\alpha \explsfe{\gamma}{\alpha}$.
This is the general pattern for simplification of the conditions
of explanation where $n=2$, $\Phi=\{\alpha\}$, $\Phi_1=\{\beta_1\}$ and
$\Phi_2=\{\beta_2\}$. Generalizing this example produces naturally the
general pattern for simplification of the conditions. 

\subsection{A formal system for inferring explanations}
\label{proofsystem}

The above ideas are embedded in a short proof system extending classical logic:

\begin{enumerate}
\item\label{propcauschema0} \textbf{\emph{Causal formulas}} \hspace{10mm}
     $(\alpha \cause \beta) \rightarrow (\alpha \rightarrow \beta)$.
\item\label{propontschema0} \textbf{\emph{Ontological atoms}}
  \begin{enumerate}
  \item\label{ontoimplicschema0}
     If $\beta \isa \gamma$ then $\beta \rightarrow \gamma$.
  \item\label{ontotranschema0}
     If $\alpha \isa \beta$ and $\beta \isa \gamma$ then $\alpha \isa \gamma$.
  \item\label{ontorefschema0}
     $\alpha \isa \alpha$\\\mbox{~}
  \end{enumerate}
\item\label{explicprop5} \textbf{\emph{Explanation atoms}}
  \begin{enumerate}
 \item\label{explicontdnupprop5} {\em Base case}\\
  $\begin{array}[t]{ll}
   \mbox{If} & \delta \isa  \beta, \;\;\delta \isa \gamma, \mbox{ and } \;
               W \not\models \neg (\alpha \wedge \delta),\\
   \mbox{then} & (\alpha \cause \beta) \quad \rightarrow \quad
        \alpha \explicc{\gamma}{\{\alpha,\delta\}}
      \end{array}$
  \item\label{transexplicprop5} {\em Transitivity (gathering the
      conditions)}\hfill 
      If\hfill $W \not \models \neg \bigwedge (\Phi \cup\Psi)$, then\\ 
       $(\alpha \explgf{\beta}{\Phi} \;\;\wedge \;\;  
                  \beta \explgf{\gamma}{\Psi})\\  
         \rightarrow \;\; \alpha \explgf{\gamma}{(\Phi \cup \Psi)}.$
  \item\label{simplexplicprop5} {\em Simplification of the set of conditions}\hfill
      If \hfill 
        $W \models \bigwedge\Phi \rightarrow \bigvee_{i=1}^n \bigwedge\Phi_i$,\\
        \makebox[0mm]{} \hfill then \hfill
       $\bigwedge_{i \in \{1,\cdots,n\}}
            \alpha \explgf{\beta}{(\Phi_i \cup \Phi)}$\\
        \makebox[0mm]{} \hfill   $\rightarrow\quad 
          \alpha \explgf{\beta}{\Phi}.$
  \end{enumerate}
\end{enumerate}

These schemes allow us to obtain the inference patterns described in the
previous section:

\noindent The base case \S~\ref{subbasecase}:
apply (\ref{ontorefschema0}) upon
(\ref{explicontdnupprop5}) where $\beta=\gamma=\delta$
prior to simplifying by means of (\ref{simplexplicprop5}).

\noindent The upward case \S~\ref{subexplicup}: apply
(\ref{ontorefschema0}) upon (\ref{explicontdnupprop5})  where $\beta=\delta$,
prior to using (\ref{simplexplicprop5}).

\noindent The downward case \S~\ref{subexplicdn}: apply
(\ref{ontorefschema0}) upon (\ref{explicontdnupprop5})  where
$\delta=\gamma$.\\ 

A more substantial application is:
\hfill $\quad C = \{\alpha \cause \beta,\;\; \gamma \cause \epsilon\},$\\
\makebox[0mm]{} \hfill $O = \{\beta \isa \gamma \},\quad
W = \{\alpha \rightarrow \beta,\;\;\beta \rightarrow  \gamma,\;\;
      \gamma \rightarrow \epsilon\}$

The first form of transitivity in Subsection~\ref{subtrans} requires that we infer:
\[\alpha \explgf{\epsilon}{\{\alpha\}}\]
Let us proceed step by step:
\[\begin{array}{cl}
\alpha \explgf{\gamma}{\{\alpha\}} & \mbox{by (\ref{explicontdnupprop5} with
  $\beta=\delta$) as upward case} \\
\gamma \explgf{\epsilon}{\{\gamma\}} & \mbox{by (\ref{explicontdnupprop5}) as base case} \\
\alpha \explgf{\epsilon}{\{\alpha,\gamma\}} & \mbox{by (\ref{transexplicprop5})} \\
\alpha \explgf{\epsilon}{\{\alpha\}} & \mbox{by (\ref{simplexplicprop5}) simplifying the proviso.}
\end{array}\]

\subsection{A generic diagram }\label{subgenerex}

\begin{figure}[h!]
\includegraphics{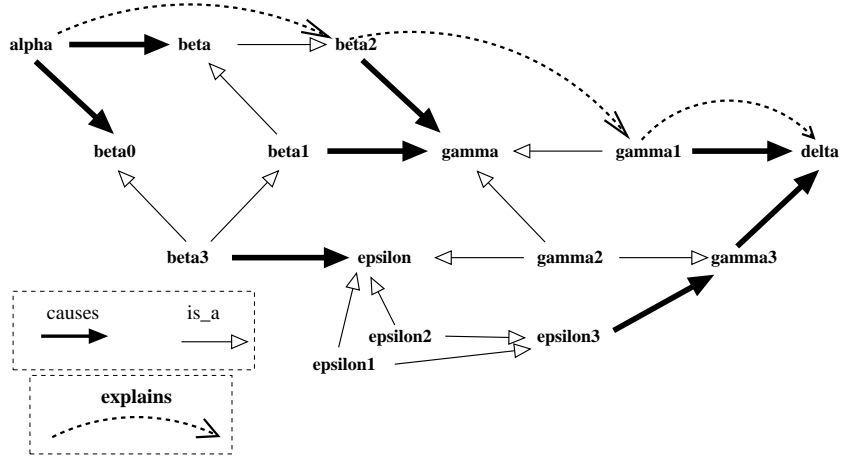}
\caption{A generic diagram, the theory with a first explaining path}  
\label{fig1}  
\end{figure}

Below an abstract diagram is depicted
that summarizes many patterns of inferred explanations from various cases of
causal statements and $\isa$ links.
The theory is described as follows (see Figure~\ref{fig1}):

\noindent$\alpha \cause \beta$, \hfill 
$\alpha \cause \beta_0$, \hfill 
$\beta_2 \cause \gamma$, \hfill 
$\beta_1 \cause \gamma$, \hfill $\,$\\
$\beta_3 \cause \epsilon$, \hfill 
$\gamma_1 \cause \delta$,  \hfill 
$\gamma_3 \cause \delta$, \hfill $\epsilon_3 \cause \gamma_3$; \hfill $\,$\\ 
$\beta \isa \beta_2$, \hfill $\beta_1 \isa \beta$, \hfill  
$\beta_3 \isa \beta_0$, \hfill $\beta_3 \isa \beta_1$, \hfill $\,$\\
$\gamma_1 \isa \gamma$, \hfill $\gamma_2 \isa \gamma$, \hfill 
$\gamma_2 \isa \gamma_3$, \hfill $\gamma_2 \isa \epsilon$, \hfill $\,$\\
$\epsilon_1 \isa \epsilon$, \hfill $\epsilon_2 \isa \epsilon$, \hfill 
$\epsilon_1 \isa \epsilon_3$, \hfill $\epsilon_2\isa \epsilon_3$. \hfill $\,$\\

\begin{figure}[h]
\includegraphics{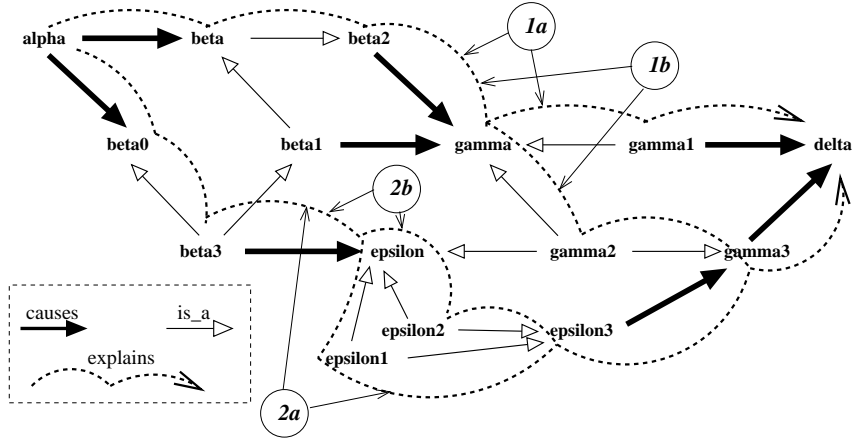}
\caption{Four optimal explanation paths from ``alpha'' to ``delta''}  
\label{fig2} 
\end{figure}

This example shows various different ``explaining paths'' from a few given
causal and ontological atoms. 
Here there is a first ``explaining path'' from $\alpha$ to $\delta$
(Figure~\ref{fig1}, see also path $(1a)$ on Figure~\ref{fig2}). 
We get successively:\hfill
$\alpha \explsf{\beta_2}{\{\alpha\}}$,\hfill \makebox[0mm]{}\\
\makebox[0mm]{}$ \hfill \alpha \explsf{\gamma_1}{\{\alpha,\gamma_1\}}$,
\hfill and \hfill $\alpha \explsf{\delta}{\{\alpha,\gamma_1\}}.$\hfill $\,$

\noindent As another ``explaining path'', we get:\hfill
$\alpha \explsf{\beta_1}{\{\alpha, \beta_1\}}$\hfill \makebox[0mm]{}\\
\makebox[0mm]{} \hfill $\alpha \explsf{\gamma_1}{\{\alpha,\beta_1,\gamma_1\}}$,
\hfill and \hfill $\alpha \explsf{\delta}{\{\alpha,\beta_1,\gamma_1\}}.$
\hfill \makebox[0mm]{}

This second path is clearly not ``optimal'' since
$\{\alpha,\gamma_1\}$ is strictly included in $\{\alpha,\beta_1,\gamma_1\}$.
The simplifying rule produces $\alpha\explsf{\delta}{\{\alpha,\gamma_1\}}$ 
from\\
 $\alpha \explsf{\delta}{\{\alpha,\beta_1,\gamma_1\}}$ but, from a
  computational point of view, it is better not to generate the second path at
  all.

Here there 
are the four ``optimal'' explanation atoms from $\alpha$ to $\delta$
(see Figure~\ref{fig2} for the precise paths):

\noindent$\!\!\begin{array}[t]{lclc}%
(1a)\!&\!\alpha \explsf{\delta}{\{\alpha, \gamma_1\}}& 
(1b)\! &\!\alpha \explsf{\delta}{\{\alpha, \gamma_2\}}\\
(2a)\!&\!\alpha \explsf{\delta}{\{\alpha, \beta_3, \epsilon_1\}}& 
(2b)\!&\!\alpha \explsf{\delta}{\{\alpha, \beta_3, \epsilon_2\}}.  
\end{array}$
\\

We have implemented a program in DLV \cite{LPFEGPS06}
(an implementation of the Answer Set Programming (known as ASP) formalism \cite{Bar03})
that takes only a few seconds to give all the results $\sigma1 \explsf{\sigma2}{\Phi}$,
for all examples of this kind, including as here when different explanation 
paths exist (less than one second for the theory depicted in
Figures~\ref{fig1} and \ref{fig2}). See \cite{Moi07} for details about this 
implementation. Let us just notice that for now,
 it does not make full simplifications: 
it tries to avoid generating explanations which are not simplified, 
in a way which is optimal in simple cases like this example.

\section{Introducing predicates}\label{secpred}
\subsection{Motivations}
In order to keep things as simple as possible, we have considered
propositional symbols only. However, we have seen various examples where
this is not really appropriate.
Indeed, stating $Heard\_loud\_bell \isa Heard\_bell$ is neither natural nor
convenient. Indeed, we should also state $Activating\_loud\_bell \isa Activating\_bell$
if necessary and so on.
It is clear that it is much more natural to state
$loud\_bell \isa bell$, and to infer the results 
about $Heard\_[...]$, $Activating\_[...]$ and so on.

We introduce predicate symbols (such as $Heard$,
$Activating$, $Own$, ...) and constant symbols such as 
$bell, noise, loud\_bell, student, book, ...$.
 
The elementary terms (represented by $\alpha, \beta$ as in preceding
sections) are ground atoms such as $Heard(bell)$, $On(alarm)$,
$Own(student,book)$. 
We need a way to use the ontological information together with the causal
information.
The ontological links concern constant symbols, as in e. g.
$a \isa b$.

We need a way to infer our old 
$Heard\_loud\_bell \isa Heard\_bell$ from the new ontological information
stating here $loud\_bell \isa bell$.
Again, since we want to keep things simple, we put some restrictions.
We distinguish two kinds of behavior for a given parameter in a predicate
(this is the main improvement from \cite{BCM07}).

Let us suppose that 
$Heard(bell)$ means ``I have heard some bell''. Then, we can say that
$Heard$ is \textbf{\em essentially existential} ({\em there exists} some bell
that I have heard). A more explicit way to express this is to
denote this predicate by $Heard_{one}$ instead of $Heard$. 
Similarly, let us suppose that $Own(student,book)$ means that ``every student
owns some book''.
We will say that the predicates $Heard$ and $Own$
\textbf{\em inherit upward} for the parameter $t$ in 
$Heard(t)$ and in $Own(t1,t)$ through the $IS-A$ hierarchy.

The other particular case of predicates, which 
\textbf{\em inherit downward} [for a given parameter]
through the $IS-A$ hierarchy, 
are \textbf{\em essentially universal} [for this parameter].
Let us take $Like$ as an example, considering that 
$Like(bell)$ means I like bells (in general: I like {\em all} the bells).
This predicate could be denoted by $Like_{all}$.
This notation allows to use
the two predicates  $Like_{one}$ and $Like_{all}$ together, 
if necessary, and has the advantage of indicating the kind of inheritance of a
predicate in its denomination.
Similarly, $Own$ inherits downward for the parameter $t1$ in $Own(t1,t)$
and could be denoted by $Own_{all,one}$.

This problem of the way predicates should inherit through the $IS-A$ hierarchy,
is a matter of formalization of natural language.
The way used here allows to avoid explicit quantifiers such as $\exists$
and $\forall$. When introducing a predicate, we must state [for each of its
parameters] whether it 
is essentially existential (then it inherits upward such as $Heard_{one}$)
or essentially universal (then it inherits downward such as $Like_{all}$), 
if we want to take advantage of the ontological
information with respect to this predicate.
Predicates which are neither ``essentially existential'' nor ``essentially
universal'' for some of their parameters cannot exploit the $IS-A$ hierarchy
for these parameters.

Let us provide the formal description of the new system.

\subsection{The vocabulary with ``upward and downward inheriting'' predicates}
\label{subvocpred}
\begin{enumerate}
\item\label{formpredvoc} \textbf{\emph{Classical vocabulary and formulas}} 
  \begin{enumerate}
  \item\label{formpredvoc1}  \textbf{\emph{Predicate symbols}}. Any arity is
    possible and, for each of their parameters, predicates can be
    \emph{essentially existential} or \emph{essentially universal} or without
    precision. The names of the predicate symbols begin with an uppercase
    letter such as in $P, Heard, On$.   
  \item\label{formpredvoc3} \textbf{\emph{Constant symbols}}. Their names begin
    with a lower case letter such as in $a, bell, loud\_bell$.
  \item\label{formpredvoc2} \textbf{\emph{Classical atoms}}. A classical atom
    is  a ground atom
    $P(a_1,...,a_n)$ where $P$ is a predicate of arity $n$ and $a_i$'s are
    constants. A propositional symbol (i.e. a predicate of arity 0)
    $P_0$ is thus a classical atom.\\ 
    Classical atoms are denoted either as $P(a_1,...,a_n)$ or $P_1(a)$ or by
    Greek letters such as $\alpha$ or $\beta_1$. 
  \item \textbf{\emph{Classical formulas}}. A classical formula is a sentence
    (Boolean combinations of classical atoms: only ground formulas are
    considered). 
  \end{enumerate}
\item\label{formpredcaus} \textbf{\emph{Causal atoms and formulas}} 
  \begin{enumerate}
  \item\label{formpredcaus0}
  If $\alpha$ and $\beta$ are classical atoms, then
  $\alpha \cause \beta$ is a \textbf{\emph{causal atom}}.
  \item\label{formpredcaus1} A  \textbf{\emph{causal formula}} is a Boolean
    combination of classical or causal atoms.
  \end{enumerate}
 \item \label{formpredont} \textbf{\emph{Ontological atoms}}\\ 
  If $a$ and $b$ are constant symbols, then 
  $a \isa b$ is an ontological atom.%
\\
  Since we want our ``predicate'' system to encompass the preceding
  ``propositional'' system, we must also consider 
  ontological links between two propositional symbols:
  if $P_0$ and $Q_0$ are two propositional symbols, then 
  $P_0 \isa Q_0$ is an ontological atom.

\item\label{formpredwco}  A \textbf{\emph{causal theory}}
 consists of a set $W$ of classical
  formulas, a set $C$  of causal formulas and a set $O$ of ontological atoms. 

\item\label{formpredexpl}  \textbf{\emph{Explanation atoms}} 
 From a given causal theory, some {\em explanation atoms} will be derived.
An {\em explanation atom} is \[\alpha \explicc{\beta}{\Phi}\]
where $\alpha$ and $\beta$ are classical atoms and $\Phi$ is a set of
classical atoms. It reads
``$\alpha$ explains $\beta$ because the set $\Phi$ is possible''.
\end{enumerate}

Basically, the derivation of the explanation atoms is as given in the
propositional case, thus now we can introduce
directly the \textbf{\em formal proof system}.

\subsection{Formal proof system of the formalism with predicates}\label{subproofpred}
\begin{enumerate}
  \item\label{predcauschema}
   \textbf{\emph{Property of the causal atoms: entailing implication}}   
  \begin{enumerate}
    \item\label{predplschemaO}
     $(\alpha \cause \beta) \rightarrow (\alpha \rightarrow \beta)$\\\mbox{~}
  \end{enumerate}
  \item\label{predontschema} \textbf{\emph{Ontological atoms}} \\
   The easiest way to present the rules in the predicate case is to 
   augment the ontology by introducing ontological links between ground atoms:

  \begin{enumerate}
    \item\label{predplschema1}
     \textbf{\emph{Deriving an augmented ontological relation}}
    \nopagebreak
    \begin{enumerate}
      \item\label{isaux1} 
        If $\alpha=P_0$ and $\beta = Q_0$ are propositional atoms, then\\ 
         $P_0 \isaaux Q_0$ whenever $P_0 \isa Q_0$.
      \item\label{isaux2}  Let $P_{all,one,\_,\cdots}$ denote some predicate of
        arity $n$, essentially universal with respect to its first parameter
        and essentially existential with respect to its second parameter (other
        parameters not concerned here, clearly the first or the second 
        parameter could similarly by the $i^{th}$ parameter for any $i \in
        \{1,\cdots,n\}$). \hfill Then\\ 
        $P_{all,one,\_,\cdots}(a_1,b_2,a_3,\cdots,a_n) \isaaux
        P_{one}(a_1,a_2,a_3,\cdots,a_n)$ whenever $a_2 \isa b_2$,
        \hfill and\\  
        $P_{all,one,\_,\cdots}(a_1,a_2,a_3,\cdots,a_n) \isaaux
        P_{one}(b_1,a_2,a_3,\cdots,a_n)$ whenever $a_1 \isa b_1$.  
    \end{enumerate}
    \item\label{predplschema2}
     \textbf{\emph{Properties of the augmented ontological relation}}
    \begin{enumerate}
      \item\label{ontotranschemaaux}\textbf{\emph{Transitivity}}\\
        If $\alpha \isaaux \beta$ and $\beta \isaaux \gamma$ then
        $\alpha \isaaux \gamma$.  
      \item\label{ontorefschemaaux}\textbf{\emph{Reflexivity}}\hspace{2em}
        $\alpha \isaaux \alpha$. 
      \item\label{predontschemaaux}\textbf{\emph{Entailing implication}}
        \hspace{1em}  
        If $\;\alpha \isaaux \beta\;$  then $\; \alpha \rightarrow
        \beta$. 
    \end{enumerate}
  \end{enumerate}
  The only difference with the propositional case is that we must use the
  augmented ontology instead of the ontology given by the user.

  \item\label{explicprop5pred} \textbf{\emph{Deriving explanation
        atoms}}
 \begin{enumerate}
  \item\label{explicontpredgener}\textbf{\emph{Base case}}\\
     $\begin{array}{ll}\mbox{If}& \beta \isaaux \gamma, \hspace{3mm}
               \beta \isaaux \delta,
               \mbox{ and } \; W \not\models \neg (\alpha \wedge \beta)\\
               \mbox{then} & \alpha \cause \gamma \quad \rightarrow \quad
               \alpha \explicc{\delta}{\{\alpha,\beta\}}.
             \end{array}$
           \item\label{explicontpredtrans}\textbf{\emph{Transitivity of
        explanation}} cf Point \ref{transexplicprop5} in \S~\ref{proofsystem}.
 \item\label{simplexplicprop5pred5}%
     \textbf{\emph{Simplifying explanation atoms}} cf Point \ref{simplexplicprop5} in \S~\ref{proofsystem}.
   \end{enumerate}
\end{enumerate}   

 Notice that we keep transitivity and reflexivity of the ontology, for the
 augmented relation $\isaaux$.
As for the ontology relation $\isa$, which is the relation  provided by the
user, we could add these two properties if desired: this would not modify the
explanation atoms.   

Here there is an example, illustrating also the interest of the general
generating rule of explanations, with ``down then up'' ontological links
(Point~\ref{explicontpredgener}):

$\begin{array}[t]{ll}
C1 & On(alarm) \cause Heard(warning\_signal);\\
C2 & Heard(loud\_noise) \cause Wake\_up;\\
O1 & loud\_bell \isa warning\_signal,\\
O2 & hooter \isa warning\_signal,\\
O3 & loud\_bell \isa loud\_noise,\\
O4 & red\_flashing\_light \isa warning\_signal.
\end{array}$\\ 

$Heard$ is supposed to be essentially existential, thus it inherits upward
(for $On$, it does not matter in this example).\\ 

We get the following ``augmented'' ontological links:\\ 
\begin{tabular}[t]{cl}
$O1aug$: & $ Heard(loud\_bell) \isaaux  Heard(warning\_signal)$,\\
$O2aug$: & $ Heard(hooter) \isaaux  Heard(warning\_signal)$,\\
$O3aug$: & $ Heard(loud\_bell) \isaaux  Heard(loud\_noise)$,\\
$O4aug$: & $Heard(red\_flashing\_light) \isaaux Heard(warning\_signal)$.
\end{tabular}\\

Thus, we get the following explanation atoms:\\
\begin{tabular}[t]{cl}
$E1$: & $On(alarm) \;\;\textit{explains}\;\; {Heard(loud\_noise)}$\\
&\multicolumn{1}{r}{\textit{bec\_poss} ${\{On(alarm),Heard(loud\_bell)\}}$}\\
$E2$: & $Heard(loud\_noise)\explsf{Wake\_up}{\{Heard(loud\_noise)\}}$\\
\multicolumn{2}{l}{$E1$ comes from $C1, O1aug$ and $O3aug$, and}\\
\multicolumn{2}{l}{$E2$ from $C2$, by the base case of explanations.}
\end{tabular}\\

Then we get, by transitivity of explanations on $E1$ and $E2$, 

$On(alarm)\ \  explains \ \ Wake\_up\ \ bec\_possible\
  \{On(alarm)$,\\\makebox[0em]{}\hfill
  $Heard(loud\_bell),Heard(loud\_noise)\}$
 and finally\\
$On(alarm)\explsf{Wake\_up}{\{On(alarm),Heard(loud\_bell)\}}$
\\
by simplifying the set of conditions, taking into account that we get\\
 $Heard(loud\_bell) \rightarrow Heard(loud\_noise)$ from $O3aug$.
\\

As a formal example, let us take the predicate variant of the example ending 
\S~\ref{proofsystem}, where $P$ denotes a unary predicate which is essentially
existential for its parameter and $\gamma$ some arbitrary classical atom.

$C =\{P(a) \cause P(b), \;\;P(c) \cause \gamma\}$,\\
\indent $O = \{b \isa c\}$,\\
\indent $W = \{P(a) \rightarrow P(b), \;\; P(b) \rightarrow P(c),\;\;
P(c) \rightarrow \gamma\}$

As in the example ending \S~\ref{proofsystem}, we get:
\[P(a) \explgf{\gamma}{\{P(a)\}}.\]
Again, let us proceed step by step:\\
$\begin{array}{cl}
P(b) \isaaux P(c) & \mbox{by (\ref{isaux2})}\\
P(a) \explgf{P(c)}{\{P(a)\}} & \mbox{by (\ref{explicontdnupprop5}) as upward case} \\
P(c) \explgf{\gamma}{\{P(c)\}} & \mbox{by (\ref{explicontdnupprop5}) as base case} \\
P(a) \explgf{\gamma}{\{P(a),P(c)\}} & \mbox{by (\ref{transexplicprop5})} \\
P(a) \explgf{\gamma}{\{P(a)\}} & \mbox{by (\ref{simplexplicprop5}) simplifying the proviso}
\end{array}$\\

As an example of a predicate of arity greater than 1, let us define a binary predicate $Own$ where
\begin{center}
 $Own(student,book)$ is intended to mean ``every student owns a book''. 
\end{center}
Let us suppose that our ontology contains
\hfill
$mary \isa student$, \hfill \makebox[0mm]{}\\
\indent  $student \isa human$ \hspace{1em} and \hspace{1em} 
$book \isa written\_document$. 

Notice that we allow ``reification'' in our formalism:
concepts such as ``student'', ``human'' and ``book'' 
are represented by constants, exactly as are
``individuals'' such as ``Mary''.
Since $Own$ is intended to mean here ``owns a'' and not ``owns all'', 
this binary
predicate is essentially existential (thus it inherits upward)
 with respect to its second parameter.
The case of the first parameter has been settled also since here 
$Own(student,book)$ means ``every student owns a book'':
$Own$ is essentially universal with respect to its first parameter.
If we need also another predicate $Own'$ where
$Own'(student,book)$ means {\em there exists a student owning a book},
it is more convenient to denote
$Own$ by $Own_{all,one}$ and $Own'$ by $Own_{one,one}$.
\\

We must state explicitly whether a predicate is essentially existential
(``${one}$'' kind)
or essentially universal (``${all}$'' kind)
(or none of these two options by default), 
with respect to each of its parameters.
This kind of problem occurs each time we want to formalize natural language:
the  user must be aware that it is important to make the
intended meaning of each predicate precise.

It is an interesting feature of our formalism that this precise meaning can be
expressed in a natural way (at least if this predicate can be used 
from ontological atoms).
So, it is not enough to mention the arity of a predicate,  its 
``${one}$'' or ``${all}$'' kind should be given for each of its parameters.

This will indicate to the system, for each parameter of a predicate, 
whether the inheritance with respect to the ontology is  ``upward''
(``${one}$'' kind parameter)
or ``downward'' (``${all}$'' kind parameter).
It is possible to use parameters for which neither the ``${one}$'' kind
nor the ``${all}$'' kind applies.
Let us consider such a predicate $P_{all, na, one}$ (``na'' for ``not
available''). Then, no augmented ontological link exists between
atoms of this predicate
where this second parameter has different values on the left side and on the
right side: if $P(t_1,t_2,t_3) \isaaux P(t'_1,t'_2,t'_3)$ is produced, then $t_2=t'_2$.\\

We would get here:

\noindent%
$\begin{array}[t]{ll}
Own_{all,one}(human,book) \;\isaaux \;Own_{all,one}(mary,book),\\
Own_{all,one}(human,written\_document) \;\isaaux\\\makebox[0em]{}\hfill%
Own_{all,one}(mary,written\_document),\\ 
Own_{all,one}(mary,book) \;\isaaux \;Own_{all,one}(mary,written\_document),\\
Own_{all,one}(human,book) \;\isaaux \\\makebox[0em]{}\hfill%
Own_{all,one}(human,written\_document).
\end{array}$

Thus, we would also get by transitivity of $\isaaux$:

\noindent
$Own_{all,one}(human,book) \isaaux  Own_{all,one}(mary,written\_document).$\\

\subsection{Extending the formalism:  ontological links between predicates}\label{subextontpred}
We could even extend the formalism so that it allows ontological links 
between predicates of arity $1$ or more.
As an example, let us suppose that we have the two unary predicates 
(of the ``one'' kind, but it does not really matter here)
$Heard$ and $Perceived$.
It is natural to state
\hspace{1em}$Heard \isa Perceived$.

We would add the following to Point~\ref{predontschema} 
\S~\ref{subproofpred} (and similarly for higher arities):

  \begin{enumerate}
\item[(${\ref{predontschema}_{ext}}'$)]\label{predontschema0predisa}
     \textbf{\emph{Ontological atoms: introducing an
      augmented relation from an ontology between predicates}}\\
If $P,Q$ are  unary predicates, then\\
     if $P \isa Q$ then
     $P(a) \isaaux Q(a)$.
\end{enumerate}

In our example, we would get
$Heard(bell) \isaaux Perceived(bell)$, $Heard(noise) \isaaux Perceived(noise)$.

In this way, it is easier and more natural to express various relations
between ``events''.
Notice that these general ontological links between predicate symbols
generalize in a natural way the ontological links given above 
(Point \ref{isaux1} in \S~\ref{subproofpred}) for propositional symbols.
 
It can be noticed that, with the ASP translation evoked
above, predicates and constants are represented by ASP constants anyway,
thus it is not harder to include the ontology between predicates.

Here there is the last example of what can be expected from this formalism:

\begin{nonumberedmath}{Example.} {\em
Getting cold usually causes Mary to become active. 
I see Mary jogging.
So, Mary getting cold might be taken as an explanation for her jogging.

$C = \{ Getting\_cold(mary) \cause Moving\_up(mary)\ \}$\\
\indent$O = \{Jogging \isa Moving\_up\}.$

For now, $W$ does not contain any special information
(only the consequences of the preceding causal and ontological atoms).

We get
$Jogging(Mary) \isaaux Moving\_up(Mary)$ from
(${\ref{predontschema}_{ext}}'$).

Thus $W= \begin{array}[t]{cl}
\{ & Getting\_cold(mary) \rightarrow Moving\_up(mary),\\
& Jogging(mary) \rightarrow Moving\_up(mary).\}
\end{array}$\\

``Mary getting cold'' can be inferred as an explanation for 
``Mary is jogging''.\\
Indeed, the causal theory entails\hfill
$Getting\_cold(mary) \textit{explains} \; {Jogging(mary)}\\\makebox[0em]{}\hfill
\textit{bec\_poss}
 \{Getting\_cold(mary), Jogging(mary)\}$.  $\;(EXPL)$

If now we add the fact that if the weather is
not cold, then Mary cannot get cold, this explanation is no longer possible
in warm weather.

Adding  the following formulas to $W$ takes the new information into account:

$Warm\_Weather$, \hspace{2em} $\neg(Warm\_Weather \wedge Cold\_Weather)$,\\
\indent $\neg Cold\_Weather \rightarrow \neg Getting\_cold(mary)$.

Then, the causal theory fails to entail the explanation atom $(EXPL)$.}
\end{nonumberedmath}

\section{About a few features of the formalism}

The explanation inference follows the patterns presented before. 
The inference pattern~(\ref{explicontdnupprop5}) in \S\ \ref{proofsystem} 
(or pattern~(\ref{explicontpredgener}) in \S~\ref{subproofpred}) is important. 

\begin{enumerate}
\item\label{justifcassis} In this inference pattern, the direction 
of the $\isa$ links $a \isa b$ and $a \isa c$ is important.
Unexpected conclusions would ensue if other directions,
e.g., $b \isa a$ and $c \isa a$, were allowed.

\item\label{justifrestr} Also, there are good reasons for 
excluding 
conclusions not endorsed by this pattern. 
One reason is that it is better to limit the explanations to a minimum,
otherwise an overwhelming set of ``explanations'' could result. 
As an example of conclusions not endorsed, notice that
no explanation atom can be derived if it does not start with 
a ground atom 
which occurs somewhere on the left side of a causal atom.
\end{enumerate}

As a short justification of these two points, let us
introduce an example which comes from a real-world
application \cite{Cetal98,BC99}, but has been drastically simplified and
reduced.  
The causal model describes a
physical system in which a {\em sliding of the flywheel} (to be abbreviated as
{\em SOF})
causes a {\em step} in the vibration measurement signal.
It is also known that {\em step} and {\em slow increase} are two kinds of
evolution of the vibration measurement signal and that a {\em sharp step} is
itself a kind of step.

This is to be formalized as
follows (the example has been  simplified for the sake of conciseness
and clarity, thus the propositional version of \S~\ref{secprop} suffices).\\ 

        (C1) $SOF \cause Step$; \\
\indent (O1) $Step \isa Evolution$, \hspace{1em}  
             (O2) $Slow\_increase \isa Evolution$\\
\indent (O3) $Sharp\_step \isa Step$.

Here are the explanation atoms which can be derived:

        $(E1)$ $SOF \explsfe{Step}{SOF}$,\\
\indent $(E2)$ $SOF \explsfe{Evolution}{SOF}$,\\
\indent $(E3)$ $SOF \explsfe{Sharp\_step}{SOF,Sharp\_step}$.\\

Let us consider Point~\ref{justifcassis} above: Our concern here is about\\
 $SOF \explsfe{Sharp\_step}{SOF,Sharp\_step}$
which can be inferred versus\\
$SOF \explsfe{Slow\_increase}{SOF,Slow\_increase}$ which cannot.

Why is it sensible to explain ``evolution'' and ``sharp step'' by
a ``sliding of the flywheel'' while ``slow increase'' could not be explained
by the same ``sliding of the flywheel''?
The reason is that, from the facts given here, a ``slow increase'' is not a
``step'' (which is indeed explained by some ``sliding of the flywheel''), 
but another kind of ``evolution''. So, explaining
``slow increase'' by some ``sliding of the flywheel'' would be unmotivated
from what we know about the system.
On the other hand, it is possible that some ``sharp step'', which is a kind of
``step'', has been provoked by some ``sliding of the flywheel''.
Notice that if there were reasons to eliminate this possibility, it should have
been noticed.
We could e.g. have added the information 
$\neg(SOF \wedge Sharp\_step)$ in $W$.
Another way would be to replace $C1$ by\hfill
 $(C1')$ $SOF \cause Moderate\_step$,\hfill $\,$\\ 
and to modify the ontological information accordingly.

This example also shows that we must exercise some care while 
stating the ontological information, which was to be expected since the
ontological information plays a great r\^{o}le in the formalism.
Since a ``step'' is an ``evolution'', there are good reasons to
explain ``evolution'' also by some ``sliding of the flywheel'':
If we have enough information to know that we get ``evolution'', but not enough
to know whether it is a ``step'' or not, it is natural to provide
``SOF'' as an explanation for this ``evolution''.
\\

Let us consider Point~\ref{justifrestr} now. For this purpose, let us add the
following information about the system: 
``Any {\em evolution} in the vibration measurement signal causes
 an {\em alarm} to be displayed on the operator control screen.''

We would then add the following formula to our causal theory.\\

$(C2)$ $Evolution \cause Alarm$.

The theory is now described by $C1$, $C2$ together with $O1$, $O2$ and
$O3$ ($W$ does not contain special information, only the consequences of
these formulas).\\

Then, the explanation atoms derived by the new theory are $E1$, $E2$,
$E3$ and $E4$, plus $E2'$ obtained by transitivity from $E2$
and $E4$ and by an obvious simplification, with $E4$ and $E2'$ as follows:

$(E4)$ $Evolution \explsfe{Alarm}{Evolution}$,\\
\indent$(E2')$ $SOF \explsfe{Alarm}{SOF}$.\\

We do not get\hspace{3mm} 
$(notE2)$ $Step \explsfe{Alarm}{Step}$.\\

The reason is that we restrict our ``explanations'' to those starting from
a ground atom 
which actually ``causes'' something, 
and $Step$ does not appear on the left side of a {\em causal atom}.
Notice however that, if $Step$ is established, then so is 
$Evolution$ from $O1$, thus, from  $E4$ we indeed get some ``explanation''
in which the left-hand side is established while the right-hand side is
 $Alarm$.

It must be noticed that here are cases 
(particularly when making abduction from a set of atoms)
where it is convenient to derive also ``explanations'' starting from
atoms such as $Step$ here. This addition is immediate in our formalism.

\section{Conclusion}

We have provided a logical framework allowing predictive and abductive
reasoning from \emph{causal} information.
Indeed, the formalism allows to express causal information in a direct way.
Then, we deduce so-called \emph{explanation atoms} which capture what 
might explain what, in view of the given information.
We have resorted to \emph{ontological} information, which
is key in generating sensible explanations from causal statements.

The user provides taxonomic information as a list of \emph{ontological atoms}
$a \isa b$ intended to mean that object $a$ ``is a'' $b$.
The basic ground atoms, 
denoted $\alpha,\beta$, are then built 
with predicates, such as $P(a,b)$.
The user provides causal information as causal atoms 
$\alpha \cause \beta$ (which can occur in more complex formulas).
This makes formalization fairly short and natural.
The ontology is used in various patterns of inference for explanations. 
Such information is easy to express, or to obtain in 
practice, due to existing ontologies and ontological languages.
If we were in a purely propositional setting, the user should write \\
 $Own\_small\_car \isa Own\_car$, $Own\_big\_car \isa Own\_car$, and also 
$Heard\_small\_car \isa Heard\_car$ and so on.

This would be cumbersome. 
In contrast, our setting is ``essentially propositional'' for what concerns
the causal atoms, in that it is 
as if $Own(small\_car)$ were a propositional symbol $Own\_small\_car$, 
while, for what concerns the ontology, we really use the fact that 
$Heard$ and $Own$ are predicates.

The notion of predicates ``essentially existential'' or ``universal'' 
allows to keep a ``datalog'' formalism,
very closed to a propositional one, without the need for explicit quantifiers
$\forall$ {\em (for all)} or $\exists$ {\em (there exists)}.\\

The present proposal is a compromise between
simplicity, as well as clarity, when it comes to describing a situation, 
and efficiency and pertinence of the results provided by the formalism.

Our work differs from other approaches in the literature 
in that it strictly separates causality, ontology and explanations.
The main advantages are that information is more properly expressed and 
that our approach is compatible with various accounts of these notions, 
most notably causality.
In particular, we need no special instances of $\alpha \cause \alpha$ 
to hold (even though $\alpha \cause \alpha$ for a particular $\alpha$ can be 
explicitly asserted). Similarly, if  $\alpha$ is equivalent to $\gamma$
and $\beta \cause \alpha$ hold, this does not mean that 
$\beta \cause \gamma$ holds. This feature
contrasts with \cite{Bel06,Boc03C,GLLMT04,HP01a,HP01b,Shaf98} although 
in the context of actions such confusion is less harmful.
Some authors have already introduced notions related to 
our causal and ontological atoms. 
In \cite{Kau91,CT94}, there are ``axioms'' which can be
loosely related to our causal and ontological atoms. Our work 
investigates how  ``explanations'' are obtained from
such causal and ontological information. 
On the other hand, we have not worked here on the important subject of what
can precisely be done from these explanation atoms.
This is left for future work, since we think that the
formalization task is also a crucial one.
As for designing some plan recognition or some abductive reasoning from our
work, this is possible with simple additions over our formalism. Let us just
give one indication here: from the sets of conditions for each explanation
atom, ``best explanations'' for a given set set of $\alpha$'s can be defined.

Also as future work we should relax some of the strong restrictions 
on the notion of ``cause'' made here.
We could introduce some ranking among the causal atoms, 
in order to cope with cases such as ``smoking causes cancer''.
Then, we could introduce the temporal aspect which is important as soon as
causation is involved, by adding a special temporal parameter.

We have designed a system in answer set programming that implements 
most of the formalism introduced above.
It is restricted to predicates of arities 0 or 1
(this could be easily extended) and the simplification part is not fully
completed (this is much harder to modify, since the computation would be 
seriously more complex, but this strong simplification does not appear to be crucial).
It works for examples of reasonable size, but it should be expandable for
some real life examples, thus showing that our two main goals
have been reached (simplicity of the formalization by a user,
and efficiency of the computation).
Our present translation in Answer set programming cannot be considered as a
competitor with achieved proposals such as the ``causal calculator'' CCalc of
\cite{GLLMT04}, but it can at least 
give some indications that a real abductive system (for instance) 
can be built over our present proposal.
We can e.g. deal with the ``cooking example'' of \cite{Kau91,CT94} by adding 
only a few rules to our program.

As another future work, we should consider ontological links 
less elementary than the taxonomic relations considered in the present system.
We think however that the present system is a good basis for 
a really practical system.

\bibliography{causesHAL} 

\section*{Acknowledgement} 
The authors thank the reviewers for their helpful and constructive comments.

\end{document}